\documentclass[times,5p,twocolumn,final]{elsarticle}

\usepackage[utf8]{inputenc} % allow utf-8 input
\usepackage[T1]{fontenc}    % use 8-bit T1 fonts
\usepackage{url}            % simple URL typesetting
\usepackage{amsfonts}       % blackboard math symbols
\usepackage{nicefrac}       % compact symbols for 1/2, etc.
\usepackage{microtype}      % microtypography
\usepackage{xcolor}         % colors
\usepackage{wrapfig}
\usepackage{graphicx}
\usepackage{subfigure}
\usepackage{booktabs} % for professional tables
\usepackage{hyperref}
\usepackage{caption}
\usepackage{makecell}

\usepackage{amsmath}
\usepackage{amssymb}
\usepackage{mathtools}
\usepackage{amsthm}
\usepackage{algorithm}
\usepackage{algorithmic}

\usepackage{multirow}
\usepackage{tabularx}
\usepackage[capitalize,noabbrev]{cleveref}

\theoremstyle{plain}
\newtheorem{theorem}{Theorem}[section]
\newtheorem{proposition}[theorem]{Proposition}

\theoremstyle{definition}

\theoremstyle{remark}

\usepackage{amsmath,amsfonts,bm}

\def\eqref#1{equation~\ref{#1}}
\def\1{\bm{1}}

\def\rve{{\mathbf{e}}}

\def\rvg{{\mathbf{g}}}

\def\rvr{{\mathbf{r}}}

\DeclareMathAlphabet{\mathsfit}{\encodingdefault}{\sfdefault}{m}{sl}
\SetMathAlphabet{\mathsfit}{bold}{\encodingdefault}{\sfdefault}{bx}{n}

\def\gB{{\mathcal{B}}}

\def\gE{{\mathcal{E}}}

\def\gG{{\mathcal{G}}}

\def\gM{{\mathcal{M}}}

\def\gO{{\mathcal{O}}}

\def\gR{{\mathcal{R}}}
\def\gS{{\mathcal{S}}}

\journal{Knowledge-Based Systems}

\begin{document}

\begin{frontmatter}	
\title{Distributed Representations of Entities in Open-World Knowledge Graphs}

\author[a,b,c]{Lingbing Guo}
\author[a,b]{Zhuo Chen}
\author[d]{Jiaoyan Chen}
\author[a,b]{Yichi Zhang}
\author[e]{Zequn Sun}
\author[f]{Zhongpu Bo}
\author[a,b,c]{Yin Fang}
\author[g]{Xiaoze Liu}
\author[a,b,c]{Huajun Chen} 
\author[h,b]{Wen Zhang \corref{mycorrespondingauthor}}  

\cortext[mycorrespondingauthor]{Correspondence to: zhang.wen@zju.edu.cn}

\affiliation[a]{organization={College of Computer Science and Technology, Zhejiang University},
	country={China}}
\affiliation[b]{organization={Zhejiang University - Ant Group Joint Laboratory of Knowledge Graph},
	country={China}}
\affiliation[c]{organization={Donghai Laboratory},
	country={China}}
\affiliation[d]{organization={Department of Computer Science, The University of Manchester},
	country={UK}}
\affiliation[e]{organization={Department of Computer Science and Technology, Nanjing University},
	country={China}}
\affiliation[f]{organization={Ant Group},
	country={China}}
\affiliation[g]{organization={Elmore Family School of Electrical and Computer Engineering, Purdue University},
	country={USA}}
\affiliation[h]{organization={School of Software Technology, Zhejiang University},
	country={China}}

\begin{abstract}
Graph neural network (GNN)-based methods have demonstrated remarkable performance in various knowledge graph (KG) tasks. However, most existing approaches rely on observing all entities during training, posing a challenge in real-world knowledge graphs where new entities emerge frequently. 
To address this limitation, we introduce Decentralized Attention Network (DAN). DAN leverages neighbor context as the query vector to score the neighbors of an entity, thereby distributing the entity semantics only among its neighbor embeddings. To effectively train a DAN, we introduce self-distillation, a technique that guides the network in generating desired representations. Theoretical analysis validates the effectiveness of our approach.
We implement an end-to-end framework and conduct extensive experiments to evaluate our method, showcasing competitive performance on conventional entity alignment and entity prediction tasks. Furthermore, our method significantly outperforms existing methods in open-world settings.
\end{abstract}


\end{frontmatter}

\section{Introduction}
\label{sec:intro}

Knowledge graphs (KGs) are pivotal in various data-driven applications~\cite{KGE-image,OpenEA,KG_survey,KGE_analysis,DBLP:journals/kbs/BiNZZYW22,DBLP:journals/ipm/BiNZHMZ0W23,guo2023newtoncotes,DBLP:conf/semweb/0007CGPYC21}. Recently, there has been a growing interest in learning low-dimensional representations, or embeddings, of entities and relations within KGs~\cite{KGE_analysis}. 
% This learning task shares similarities with the process of representing words in a text corpus~\cite{word2vec,word2vec_nips}, where the common goal is to encode an element’s relationship with other elements into embedding.
This endeavor, akin to word representation in text corpora~\cite{word2vec,word2vec_nips}, aims to encapsulate relationships between elements in embeddings.

The existing methods for KG embedding and word embedding exhibit even more similarities. As shown in Figure~\ref{fig:kge-we}, the KG comprises three triplets conveying similar information to the example sentence. Triplet-based KG embedding models like TransE~\cite{TransE} transform the embedding of each subject entity and its relation into a hidden vector, subsequently used to predict the central entity \textit{W3C} of the triplets. This behavior resembles that of the Skip-gram model~\cite{word2vec}, where each word embedding within a window is used to predict the central word. For graph neural network (GNN)-based KG embedding models, e.g., GCN-Align~\cite{GCN-Align} and CompGCN~\cite{CompGCN}, they aggregate the embeddings of \textit{W3C} and its neighbors to obtain its output embedding. The aggregation operation mirrors the CBOW model~\cite{word2vec}, except that CBOW does not involve self-embedding.

\begin{figure*}
	\centering
	\includegraphics[width=\textwidth]{kge_we.png}
	\caption{A comparison between KG embedding and word embedding. Left: the KG and the sentence contain the same information. Center: the triplet-based models are similar to Skip-gram where each neighbor embedding is used to predict the central element. Right: the GNN-based models resemble CBOW where all neighbor embeddings are aggregated to represent the central element, with the exception that CBOW does not uses the self-entity embedding.}
	\label{fig:kge-we}
\end{figure*}

Within the realm of encoding relational information, it becomes pertinent to question the necessity of incorporating the self-entity when aggregating neighborhood information. In this paper, we delve into this question and find that, at least concerning encoding relational information, the answer may lean towards the negative.

Consider the instance of encoding the relational information of the entity \textit{W3C} into an embedding. All relevant information is structured in the form of triplets, such as $(\textit{RDF}, \textit{developer}, \textit{W3C})$. Removing the self-entity \textit{W3C} does not compromise the integrity of the information. One might argue that \textit{W3C} carries useful information like images and attributes. However, multi-modal KG embedding methods often encode different modalities of information separately and then merge the outputs through a fusion layer~\cite{MMEA,EVA,MSNEA,MCLEA,Meaformer}. Hence, excluding the self-entity when encoding relational information appears reasonable.

Drawing inspiration from the CBOW schema, we propose Decentralized Attention Network (DAN) to distribute the relational information of an entity exclusively over its neighbors.
DAN retains complete relational information and empowers the induction of embeddings for new entities. For example, if \textit{W3C} is a new entity, its embedding can be computed based on its established relationships with existing entities, such as \textit{Tim Berners-Lee}, \textit{RDF}, and \textit{XML Schema}. In contrast, the existing methods additionally rely on the embedding of \textit{W3C}, thus constraining their capacity to generate embeddings for new entities.

Moreover, DAN introduces a distinctive attention mechanism that employs the neighbors of the central entity to evaluate the neighbors themselves. This collective voting mechanism helps mitigate bias and contributes to improved performance, even on traditional tasks. It also distinguishes DAN from other existing inductive methods that primarily focus on new or unknown entities~\cite{OOKB,VNNetwork,LAN,GraIL,NBFNet,REDGNN,GCC}.

We introduce a self-distillation module to assist DAN in autonomously learning to generate output embeddings. Take the top-right sub-figure of Figure \ref{fig:kge-we} as an example, we denote the input embedding of \textit{W3C} as $\rve_{\text{W3C}}$ (the yellow unit) and the output embedding as $\rvg_{\text{W3C}}$ (the orange unit). 
Although $\rve_{\text{W3C}}$ dose not directly contribute to its output embedding $\rvg_{\text{W3C}}$, it plays a pivotal role in learning the embeddings of its neighbors, such as $\rvg_{\text{Tim Berners-Lee}}$ and $\rvg_{\text{RDF}}$. 
Hence, $\rve_{\text{W3C}}$ may encapsulate valuable knowledge about its neighbors, which we would like to distill into $\rvg_{\text{W3C}}$ by maximizing their mutual information (MI)~\cite{MINE}. This process not only helps DAN to encode the relational information of known entities, but also trains DAN to produce desired embeddings for new entities by effectively encoding the neighbors to align with the output embedding of the known entities. Theoretical proofs are provided to substantiate this concept.

To evaluate our approach, we implement an end-to-end decentralized KG representation learning framework called \emph{decentRL} and conduct comprehensive experiments on two common KG embedding tasks: entity alignment and entity prediction. The experimental results demonstrate that decentRL achieves superior performance on benchmarks of these two tasks, both with and without new entities. The source code and datasets are available at github.com/guolingbing/Decentralized-Attention-Network.

\section{Related Works}

We category related work into two primary areas: conventional and inductive KG embedding.

\subsection{Conventional KG Embedding}
Conventional KG embedding approaches broadly fall into two types:  
% Triplet-based methods can be further divided into translational methods~\cite{TransE,MTransE,JAPE}, semantic matching methods \cite{DistMult,ComplEx,RotatE,TuckER}, and neural methods~\cite{ConvE,RSN}.
Triplet-based and GNN-based methods. Triplet-based methods include translational methods~\cite{TransE,MTransE,JAPE}, semantic matching methods~\cite{DistMult,ComplEx,RotatE,TuckER}, and neural methods~\cite{ConvE,RSN}.  For a detailed understanding, interested readers can refer to surveys~\cite{KG_survey,Survey,BootEA}. 
GNN-based methods~\cite{CompGCN,R-GCN,RDGCN,AliNet,MuGNN,Dual_graph,AVR-GCN} introduce relation-specific composition operations to combine neighbors and their corresponding relations before performing neighborhood aggregation. They usually leverage existing GNN models, such as GCN and GAT~\cite{GCNs,GAT}, to aggregate an entity’s neighbors. It is worth noting that these GNN models are regarded as inductive models in graph representation learning where nodes possess self-features. In relational KG embedding, entities do not have such features, which restricts their capacity to induce embeddings for new entities.

\subsection{Inductive KG Embedding}
The existing KG embedding methods with inductive ability can be categorized as follows:

\paragraph{Multi-modal Methods} These methods~\cite{KGE-image,EVA,MSNEA,MCLEA,Meaformer,KGE-desc,DBLP:conf/semweb/ChenGFZCPLCZ23,DBLP:journals/corr/abs-2305-14651} integrate image and attribute information to generate embeddings for unseen entities in KG embedding. 
% However, their relational encoding modules remain transductive, making them less pertinent to our work. Therefore, we do not delve into their specifics in this paper.
Their relational encoding modules, however, remain transductive and thus are not the primary focus of our study.

\paragraph{Inductive Relation Prediction Methods} Methods such as GraIL~\cite{GraIL}, along with its successors GCC~\cite{GCC}, RED-GNN~\cite{REDGNN}, INDIGO~\cite{INDIGO}, NBFNet~\cite{NBFNet} are tailored for relation prediction without generating entity embeddings.  
Hence, They are unsuitable for tasks like entity alignment~\cite{OpenEA,JAPE}. 

\paragraph{Few-shot Entity Prediction Methods} Few-shot entity prediction methods, including MetaR~\cite{MetaR} and its successor GEN~\cite{GEN}, adopt a meta-learning approach. Unlike inductive relation prediction, these methods address unseen relations. The task setting differs from conventional entity prediction, where a support triplet set specific to a relation $r$ is provided to predict the missing entities in the query set related to $r$. Each training/testing example comprises a support set and a query set.

\paragraph{Out-of-KG Entity Prediction Methods} Out-of-KG entity prediction methods, such as MEAN~\cite{OOKB}, VN Network~\cite{VNNetwork}, and LAN~\cite{LAN}, leverage logic rules to infer the missing relationships but do not generate unconditioned entity embeddings for other tasks. These methods share a similar task setting with ours, where all relations are known during training. The new entities are unknown but connected to the training KG.

In light of the above discussion, we emphasize the novelty of our method in three key aspects:

\paragraph{Applicability} Our method represents a standard KG embedding approach capable of generating embeddings for various tasks. This distinguishes it from most inductive methods that either cannot produce entity embeddings~\cite{GraIL,NBFNet,GCC}, or have entity embeddings conditioned on specific relations/entities~\cite{VNNetwork,LAN}. While some methods attempt to address entity alignment by introducing a new relation, the results often demonstrate poor performance, as evidenced in \cite{OpenEA,MTransE}. 

\paragraph{Performance} Unlike many inductive methods that are solely evaluated on datasets with unseen entities, our method aims to produce high-quality embeddings for both seen and unseen entities across various downstream tasks. To our knowledge, decentRL is the first method capable of generating high-quality embeddings for different downstream tasks on datasets that encompass both existing and new entities.

\paragraph{Generality} The proposed DAN is compatible with most existing GNN-based methods, allowing these methods to leverage our DAN as the GNN module for entity encoding. Furthermore, the computational cost is comparable to that of existing methods. Therefore, we offer an efficient and general GNN architecture for KG embedding.

\begin{figure*}[t]
	\centering
	\includegraphics[width=.95\textwidth]{dan.png}
	\caption{Insight into multi-layer DAN. a. In the single-layer DAN, we first use an additional aggregation layer to obtain the neighbor context (1-2); we then use the neighbor context as query to score neighbors (3); we finally aggregate the neighbors with the attention scores to obtain the final output embedding (4-5). b. In the multi-layer DAN, we first use the output embedding of \textit{W3C} at layer $k-1$ as query to score the output embedding of its neighbors at layer $k-2$ (1); we then aggregate the neighbor embeddings at layer $k-2$ with the attention scores to obtain the output embedding of W3C at layer $k$ (2-3); similarly, we use the output embedding of \textit{W3C} at layer $k$ as query to score the output embedding of its neighbors at layer $k-1$, and finally use the attention scores to aggregate the neighbor embeddings at layer $k-1$ to obtain the output embedding of W3C at layer $k+1$ (4-6).}
	\label{fig:networks}
%	\vspace{-1em}
\end{figure*}

\section{Decentralized Attention Network}
\label{sec:DAN}
In this section, we discuss the proposed DAN in detail. We start from preliminaries and then introduce DAN and how to use it in different tasks.

\subsection{Preliminaries}

\begin{table}[t]  
\small
\centering  
\caption{The symbols used in the paper.}
\resizebox{.95\linewidth}{!}{\scriptsize
\begin{tabular}{l|l}  
\toprule  
Symbol & Description \\  
\midrule  
$\mathcal{G}$ & the knowledge graph \\  
$\mathcal{E}$ & the entity set \\  
$\mathcal{R}$ & the relation set \\
$\mathcal{T}$ & the triplet set \\
$\mathcal{S}$ & the training alignment set \\
$e_i$ & an entity \\
$\mathbf{e}_i$ & the input embedding of entity $e_i$\\
$\mathbf{g}_i^k$ & the output embedding of entity $e_i$ at layer $k$\\
$r_i$ & a relation \\ 
$\mathbf{r}_i$ & the input embedding of relation $r_i$ \\
$\mathbf{W}_i^k$ & a weight matrix at layer $k$ \\
$\sigma$ & the activation function LeakyReLU~\cite{LeakyReLU} \\
$a_{ij}$ & the attention score from entity $e_i$ to entity $e_j$ \\
$f(\mathbf{g}, \mathbf{e})$ & the mutation information estimation function \\
$\hat{\mathbf{e}}_i$ & the copy of the input embedding for entity $e_i$ \\
$N_i$ & the neighbor set of entity $e_i$\\
$X_j$ & the negative sampled entity set for $e_j$ in self-distillation\\
$I(\mathbf{g}_i, \mathbf{e}_i)$ & the mutual information between $\mathbf{g}_i$ and $\mathbf{e}_i$\\
\bottomrule  
\end{tabular}}  
\label{tab:symbol}
\end{table}

We start by outlining fundamental concepts about knowledge graphs and listing all utilized symbols in Table~\ref{tab:symbol}.
 
\paragraph{Knowledge Graph} 
A knowledge graph (KG) is a multi-relational graph where nodes represent real-world entities and edges represent different relationships between entities. We define a KG as a 3-tuple $\mathcal{G} =(\mathcal{E},\mathcal{R},\mathcal{T})$, where $\mathcal{E}$ and $\mathcal{R}$ are sets of entities and relations, respectively. $\mathcal{T}$ represents the set of triplets in the form of (\textit{subject entity}, \textit{relation}, \textit{object entity}).

\paragraph{Entity Alignment}
Entity alignment aims to identifying the potentially aligned entity pairs in two different KGs $\mathcal{G}_1 =(\mathcal{E}_1,\mathcal{R}_1,\mathcal{T}_1)$ and $\mathcal{G}_2 =(\mathcal{E}_2,\mathcal{R}_2,\mathcal{T}_2)$, given a limited number of aligned pairs as training data $\mathcal{S} \subset \mathcal{E}_1 \times \mathcal{E}_2$. In entity alignment, $\mathcal{G}_1$ and $\mathcal{G}_2$ are often merged to a joint KG $\mathcal{G} =(\mathcal{E},\mathcal{R},\mathcal{T})$, allowing the model to learn representations in a unified space~\cite{FuAlign,DBLP:conf/acl/GuoHZC22,DBLP:conf/icml/GuoZSCHC22}. This enables entities with similar relational information in different KGs to have close embeddings in the vector space. Generally, entity alignment involves comparing the distances between output entity embeddings in their respective KGs to identify aligned entity pairs, which directly reflects the performance of a KG embedding method.

\paragraph{Entity Prediction}
Entity prediction, also known as KG completion~\cite{TransE}, aims to predict the missing subject entity $e_1$ or object entity $e_2$, given an incomplete triplet $(?,r,e_2)$ or ($e_1,r,?)$ as input. Unlike entity alignment, the performance of entity prediction also relies on the ability of the predictor~\cite{RSN}. Many methods even lack an embedding module~\cite{GraIL,REDGNN,GCC,NeuralLP,NeuralSymbolicQA}. 

\paragraph{Open-World Knowledge Graphs} In real-world KGs, the number of entities is not constant and new entities emerge frequently. To address this practical setting, we proposed open-world entity alignment, where a proportion of entities in the testing set are unseen to the model. We remove the relevant triplets from $\mathcal{T}_1$ and $\mathcal{T}_2$, setting them as auxiliary triplets which are available only at the testing stage. Similarly, for open-world entity prediction, we also remove the relevant triplets from the training set and set them as auxiliary triplets. Details on how to construct open-world datasets are introduced later in Section~\ref{sec:expr_dataset}.

\paragraph{Graph Attention Network} 
For an entity $e_i$, graph attention network (GAT) \cite{GAT} aggregates the embeddings of its neighbors $N_i$ and itself into a single output vector $\mathbf{g}_i$ as follows:
\begin{align}
	\label{eq:gat}
	\mathbf{g}_i = \sum_{e_j\in N_i\cup\{e_i\}  }{a_{ij}\mathbf{W}\mathbf{e}_j},
\end{align}
where $a_{ij}$ is the learnable attention score from $e_i$ to $e_j$, and $\mathbf{W}$ is the weight matrix. To obtain $a_{ij}$, a linear attention mechanism is used:
\begin{align}
	\label{eq:att_w}
	a_{ij} = \textit{softmax}\big(\sigma(\mathbf{a}^\mathsf{T}[\mathbf{W}_1 \mathbf{e}_i \,\rVert\, \mathbf{W}_2 \mathbf{e}_j])\big),
\end{align}
where $\mathbf{a}$ is a weight vector to convert the concatenation of two embeddings into the attention score, and $\rVert$ denotes the concatenation operation. $\mathbf{W}_1$ and $\mathbf{W}_2$ are weight matrices. $\sigma$ is the activation function LeakyReLU \cite{LeakyReLU}. 

The attention mechanism in GAT consists of two steps: weighting and aggregation. In the weighting step, attention scores are computed for each entity in linear complexity. Subsequently, in the aggregation step, a weighted average is applied to the entity embeddings. This efficient implementation makes it suitable for large-scale knowledge graphs~~\cite{RDGCN,AliNet}. 

However, GAT also has some limitations. When encountering a new entity (e.g., \textit{W3C}), its embedding $\rve_\text{W3C}$ is randomly initialized, and the computed attention scores by GAT are meaningless. Additionally, $\rve_\text{W3C}$ is also a noise vector in the aggregation step.

\subsection{Decentralized Attention Network} 

If $\mathbf{e}_\text{W3C}$ is unobservable during the training phase, it becomes less useful and potentially detrimental when computing attention scores during the testing phase. To address this issue, we can introduce a decentralized attention network. 
In the decentralized approach, all entities (including the unseen entities) are still randomly-initialized, but the attention layer requires two different types of inputs: the neighbor context vector as the query vector, and neighbor embeddings as the key and value vectors. As shown in Figure \ref{fig:networks}a, we can initially employ an independent module to aggregate neighbor embeddings and obtain the context vector, followed by attention-based weighting and aggregation steps. However, this implementation involves additional computation at each layer and requires an extra round of neighbor aggregation after obtaining the context vector, which can be cumbersome.

Alternatively, we can implement the decentralized approach using a second-order attention mechanism. As depicted in~\ref{fig:networks}b, each layer in DAN consists of two steps, similar to a multi-layer GAT. The computation involves the previous two layers and can be formulated using the following equation:  
\begin{align}
	\label{eq:dan}
	a_{ij} = \textit{softmax}\big(\sigma(\mathbf{a}^\mathsf{T}[\mathbf{W}_1 \mathbf{g}_i^\prime\rVert\, \mathbf{W}_2 \mathbf{g}_j^{''} ])\big),
\end{align}
where $\mathbf{g}_i^\prime$, $\mathbf{g}_j^{''}$ denote the output embeddings of the last layer for $e_i$ and of the penultimate layer for $e_j$, respectively. Notably, \emph{if we consider $\mathbf{g}_i^\prime$ as the neighbor aggregation of the penultimate layer for $e_i$, then $\mathbf{g}_j^{''}$ precisely denotes the embedding of $e_j$ used in summing $\mathbf{g}_i^\prime$}. Consequently, $a_{ij}$ can represent the attention score of $e_i$'s neighbor context to $e_j$. This second-order modeling can significantly reduce the cost of producing neighbor context for a multi-layer model. Subsequently, the output embedding is obtained as follows:
\begin{align}
	\label{eq:dat}
	\mathbf{g}_i = \sum_{e_j \in N_i  }{a_{ij}\mathbf{W}\mathbf{g}^{''}_j}.
\end{align}
It is important to note that we perform aggregation over the next-to-last layer because the score $a_{ij}$ is computed with the neighbor embeddings at this layer as keys. It ensures that the output embeddings are consecutive and enhances the correlation of outputs between different layers.

\begin{figure*}[t]
	\centering
	\includegraphics[width=\textwidth]{distillation.png}
	\caption{An illustration of self-distillation. The yellow and orange cells denote the input and decentralized embeddings, respectively.}
	\label{fig:distillation}
\end{figure*}

For the first layer of DAN, we initialize $\mathbf{g}_i^0$ and $\mathbf{g}_j^{-1}$ using the following equation:
\begin{align}
	\label{eq:mean_aggregator}
	\mathbf{g}_{i}^0 =  \frac{1}{|N_i|} \sum_{e_j \in N_i} \mathbf{W}^0 \mathbf{e}_j, 
	\quad \mathbf{g}_{j}^{-1} = \mathbf{e}_j.
\end{align}
We use a mean aggregator to obtain the decentralized embedding $\mathbf{g}_{i}^0$, and other pooling aggregators~\cite{Pooling} can also be employed. Therefore, DAN is as efficient as GAT, with the only divergence being the additional aggregator (Equation (\ref{eq:mean_aggregator})).

\subsection{Adaption to Different Tasks}
We employ different adaptation strategies for various KG embedding tasks. In entity alignment, we follow the existing GNN-based methods~\cite{GCN-Align,AliNet} to concatenate the output embeddings from each layer to form the final representation. This process can be written as follows:
\begin{align}
	\label{eq:concat}
	\mathbf{g}_i = [ \mathbf{g}_i^1 \,\rVert\,\ldots\,\rVert\,\mathbf{g}_i^{K} ].
\end{align}
where $\mathbf{g}_i$ represents the final representation of entity $e_i$, and $K$ denotes the number of DAN layers.

In entity prediction, we use the output of the last layer as the final embedding, which enables us to select larger batch-sizes and hidden-sizes for improved performance. This can be formulated as:
\begin{align}
	\label{eq:concat2}
	\mathbf{g}_i = \mathbf{g}_i^{K},
\end{align}

\section{Learning to Generate Decentralized Embeddings}
\label{sec:self-distillation}

In this section, we introduce self-distillation to instruct DAN in generating desired embeddings for different tasks.

\subsection{The Correlation between Input Embedding and Decentralized Embedding}

To gain a deeper understanding of self-distillation, it is essential to analyze the relationship between the input embedding and the decentralized output embedding. Let’s consider the example of the entity \textit{W3C}, denoted as $\mathbf{e}_\text{W3C}$ for the input embedding and $\mathbf{g}_\text{W3C}$ for the decentralized embedding. As \textit{W3C} has several neighbors, , such as \textit{Tim Berners-Lee}, \textit{RDF}, and \textit{XML Schema}, DAN uses the embeddings $\mathbf{e}_\text{Tim Berners-Lee}$, $\mathbf{e}_\text{RDF}$, and so forth, as input to generate $\mathbf{g}_\text{W3C}$ (Figure~\ref{fig:distillation}a).

Now, let’s consider a scenario where DAN is responsible for generating embeddings for the neighbors of \textit{W3C}, specifically $\mathbf{g}_\text{Tim Berners-Lee}$, $\mathbf{g}_\text{RDF}$, and $\mathbf{g}_\text{XML Schema}$. In this context, $\mathbf{e}_\text{W3C}$ is employed as one of the input embeddings (Figure \ref{fig:distillation}b). Consequently, if \textit{W3C} is a known entity in the training set, $\mathbf{e}_\text{W3C}$ can assimilate some information about its neighborhood through back-propagation. We anticipate that $\mathbf{g}_\text{W3C}$ should, at the very least, encapsulate the information embedded in $\mathbf{e}_\text{W3C}$ to more effectively represent \textit{W3C}. In the case where \textit{W3C} is an unknown entity, it becomes even more critical for DAN to learn how to encode this entity using its neighbor embeddings as input. 

Based on these observations, we propose self-distillation as a method to prompt DAN to distill knowledge from $\mathbf{e}_\text{W3C}$ as a cue for encoding the neighbor embeddings (Figure~\ref{fig:distillation}c).

\subsection{Self-Distillation}
In self-distillation, we employ a teacher-student framework, where the input embedding $\mathbf{e}_i$ serves as the teacher model and the corresponding decentralized embedding $\mathbf{g}_i$ acts as the student model~\cite{teacherstudent1}. The objective is for $\mathbf{g}_i$ to learn as much as possible from the teacher $\mathbf{e}_i$. Notably, the fundamental difference between self-distillation and conventional student-teacher models~\cite{teacherstudent1,teacherstudent2,CRD} lies in the fact that $\mathbf{e}_i$ is not constant in our approach. 

The objective for self-distillation can be defined as follows:
\begin{align}
	\label{eq:full_objective}
	\gO_\textit{self-distillation} = &\mathop{\text{argmax}}_{\mathbf{e}_i} \sum_{e_j\in N_i} \mathop{\mathbb{E}}_{X_j} \log\Big( \frac{f(\mathbf{g}_j, \hat{\mathbf{e}}_j)}{ \sum_{e_k \in X_j} f(\mathbf{g}_j, \hat{\mathbf{e}}_k)} \Big) \nonumber\\
	 &+\mathop{\text{argmax}}_{\mathbf{g}_i, f} \mathop{\mathbb{E}}_{X_i} \log\Big( \frac{f(\mathbf{g}_i, \hat{\mathbf{e}}_i)}{ \sum_{e_j \in X_i} f(\mathbf{g}_i, \hat{\mathbf{e}}_j)} \Big),
\end{align}
where $N_i$, $X_j$ denote the neighbor set for $e_i$ and sampled negative entity set for $e_j$, respectively. The objective consists of two terms. The first term aims to find the optimal $\rve_i$ for its neighbors $N_i$ (Figure~\ref{fig:distillation}b). The second term aims to distill knowledge from $\rve_i$ into $\rvg_i$ (Figure~\ref{fig:distillation}c). Here, $f$ represents the estimation function used to estimate the correlation score between $\rve_i$ and $\rvg_i$. We define $f: \mathbb{R}^O \otimes \mathbb{R}^I \rightarrow  \mathbb{R}$, where $O$ and $I$ denote the dimensions of the output and input embeddings, respectively. More specifically, it is calculated as follows:
\begin{align}
	\label{eq:density_function}
	f(\mathbf{g}_i, \mathbf{e}_i) = \exp( \mathbf{g}_i^\mathsf{T} \mathbf{W}_{f} \hat{\mathbf{e}}_i + \mathbf{b}_f),
\end{align}
where $\mathbf{W}_{f}$ and $\mathbf{b}_f$ represent the weight matrix and bias, respectively. $\hat{\mathbf{e}}_i$ denotes the copy of $\mathbf{e}_i$ without back-propagation. It is expected that $f(\mathbf{g}_i, \hat{\mathbf{e}}_i)$ is significantly larger than $f(\mathbf{g}_i, \hat{\mathbf{e}}_j)$ for $j\neq i$, which leads to a contrastive loss in Equation~(\ref{eq:full_objective}) with $X_i$, $X_j$ as sampled negative sets~\cite{MINE,SSL}.

The selection of mutual information density as $f$ is deliberate, as it offers a more adaptable approach to matching the two embeddings $\mathbf{e}_i$ and $\mathbf{g}_i$. In contrast to distance-based estimation methods (e.g., L1/L2 distance) that strive for precise matching at every dimension, the proposed method recognizes that the two embeddings serve different functions and occupy distinct positions in DAN. Therefore, rather than enforcing an exact match between $\mathbf{e}_i$ and $\mathbf{g}_i$, the proposed self-distillation implicitly aligns them by maximizing their mutual information $\widehat{I}(\mathbf{g}_i, \hat{\mathbf{e}}_i)$ parameterized by $f$.

\begin{table*}[!t]
	\centering
	\caption{Statistics of the datasets.}
	
	\label{tab:ea_dataset}
	\resizebox{.95\linewidth}{!}{\scriptsize
		\begin{tabular}{lcccccccc}
			\toprule 
			\multirow{2}{*}{Dataset}  & Original & \multicolumn{2}{c}{With new entities} & \multirow{2}{*}{\# Entities} & \multirow{2}{*}{\# Relations} & \multirow{2}{*}{\# \makecell{Training\\ alignments}} & \multirow{2}{*}{\#\makecell{Testing \\alignments }}&  \multirow{2}{*}{\# \makecell{Testing\\ triplets}}  \\
			\cmidrule(lr){2-2}\cmidrule(lr){3-4}  &\#Train triplets & \#Train triplets & \#Auxiliary triplets \\ 
			\midrule
			\multirow{2}{*}{{ZH-EN}} & 70,414 & 53,428 & 16,986 & 19,388 & 1,701 & 4,500 & 10,500 &  \\
			& 95,142 & 72,261 & 22,881 & 19,572 & 1,323 & 4,500 & 10,500 & -  \\
			\midrule
			
			\multirow{2}{*}{{JA-EN}} & 77,214 & 57,585 & 19,629 & 19,814 & 1,299 & 4,500 & 10,500 &  \\
			& 93,484 & 69,479 & 24,005 & 19,780 & 1,153 & 4,500 & 10,500 & - \\
			\midrule
			
			\multirow{2}{*}{{FR-EN}} & 105,998 & 79,266 & 26,732 & 19,661 & 903 & 4,500 & 10,500 &  \\
			& 115,722 & 87,030 & 28,692 & 19,993 & 1,208 & 4,500 & 10,500 & - \\
			
			\midrule
			FB15K-237 & 272,115 & 170,463 & 101,652 & 14,541 & 237 & -  &  - & 20,466\\ 
			\bottomrule
	\end{tabular}}
\end{table*}

\begin{table}[t]
	\centering
	\small
	\caption{Hyper-parameter settings in the main experiments.}
	\resizebox{\linewidth}{!}{
		\begin{tabular}{lccccc}
			\toprule
			% \multicolumn{11}{c}{\cellcolor{lightgray} GMN}\\
			% \midrule
			Dataset & \# Epoch & Batch-size &  \# Layers &  \makecell{Learning \\ rate} & \# \makecell{Negative \\samples\\per entity} \\
			\midrule
			ZH-EN & 80 & 4,500  & 4 & 0.001 & 10 \\
			JA-EN & 80 & 4,500 & 4 & 0.001  & 10 \\
			FR-EN & 80 & 4,500 & 4 & 0.001  & 10 \\
			% \midrule
			\midrule
			& \# Epoch & Batch-size &  \# Layers &  \makecell{Learning \\ rate} & Decoder \\
			\midrule
			FB15K & 500 & 2,048 & 2 & 0.003  &  ComplEX\\
			WN18 & 500 & 2,048 & 2 & 0.001  & ComplEX \\
			FB15K-237 & 500 & 2,048 & 2 & 0.001  & ComplEX \\
			WN18RR & 500 & 2,048 & 2 & 0.001  & DistMult \\
			\bottomrule
	\end{tabular}}
	\label{tab:hyperparameter}%
\end{table}%

It is worth noting that we use the copy value of $\mathbf{e}_i$ in Equation~(\ref{eq:full_objective}) to prevent the input embedding $\mathbf{e}_i$ from inadvertently aligning with $\mathbf{g}_i$ in reverse. This approach carries two significant benefits. Firstly, it enables the learning of $\mathbf{g}_i$ while optimizing the neighbors of $e_i$, and vice versa, without the necessity of alternating optimization of these two terms. Second, even with the use of the copied $\mathbf{e}_i$ without alternative training, self-distillation remains clear and well-defined. 

We provide two propositions to substantiate these claims:

\begin{proposition}[lower bound]
	\label{prop:lower-bound}
	In self-distillation, the mutual information between $\mathbf{e}_i$ and $\mathbf{g}_i$ is lower-bounded as follows:
	\begin{align}
		\widehat{I}(\mathbf{g}_i, \hat{\mathbf{e}}_i)  \leq I(\mathbf{g}_i, \mathbf{e}_i),\quad \forall e_i \in \gE
	\end{align}
\end{proposition}

\begin{proposition}[optimization]
	\label{prop:auto-distillation}
	Optimizing the first term in Equation~(\ref{eq:full_objective}) w.r.t. entity $e_i$ is equivalent to optimizing the second term in Equation~(\ref{eq:full_objective}) w.r.t., the neighbors of $e_i$:
	
	\begin{align}
		\mathop{\text{argmax}}_{\mathbf{e}_i} \sum_{e_j\in N_i} \mathop{\mathbb{E}}_{X_j} \log\Big( \frac{f(\mathbf{g}_j, \hat{\mathbf{e}}_j)}{ \sum_{e_k \in X_j} f(\mathbf{g}_j, \hat{\mathbf{e}}_k)} \Big) \\\nonumber
		\Leftrightarrow  
		\sum_{e_j\in N_i} \mathop{\text{argmax}}_{\mathbf{g}_j, f} \mathop{\mathbb{E}}_{X_j} \log\Big( \frac{f(\mathbf{g}_j, \hat{\mathbf{e}}_j)}{ \sum_{e_k \in X_j} f(\mathbf{g}_j, \hat{\mathbf{e}}_k)} \Big)
	\end{align}
	
\end{proposition}

\begin{proof}
Please refer to \ref{proof:low-bound} and \ref{proof:auto-distillation}.
\end{proof}

\begin{table*}[t]
	\centering
	\caption{The conventional entity alignment results on DBP15K datasets. The \textbf{boldfaced} and \underline{underlined} denote the best and second-best results, respectively. Average of $5$ runs, the same below.}
	\resizebox{.80\textwidth}{!}{
		\begin{tabular}{lcccccccccc}
			\toprule
			\multirow{2}{*}{Method} & \multirow{2}{*}{\makecell{Backbone\\model}} & \multicolumn{3}{c}{{ZH-EN}} & \multicolumn{3}{c}{{JA-EN}} & \multicolumn{3}{c}{{FR-EN}} \\
			\cmidrule(lr){3-5} \cmidrule(lr){6-8} \cmidrule(lr){9-11} 
			& & Hits@1 & Hits@10 & MRR & Hits@1 & Hits@10 & MRR & Hits@1 & Hits@10 & MRR \\ \midrule
			
			JAPE~\cite{JAPE}  &  None & 0.412 & 0.745 & 0.490 & 0.363 & 0.685 & 0.476 & 0.324 & 
			0.667& 0.430 \\  
			RotatE~\cite{RotatE}  & None & 0.485 & 0.788 & 0.589 & 0.442 & 0.761 & 0.550 & 0.345 & 0.738 & 0.476 \\
			\midrule
			ConvE~\cite{ConvE}  & CNN & 0.169 & 0.329 & 0.224 & 0.192 & 0.343 & 0.246 & 0.240 & 0.459 & 0.316 \\
			RSN~\cite{RSN}  & RNN & 0.508 & 0.745 & 0.591 & 0.507 & 0.737 & 0.590 & 0.516 & 0.768 & 0.605 \\
			\midrule
			GCN-Align~\cite{GCN-Align}  & GCN & 0.413& 0.744 & 0.549 & 0.399 & 0.745 & 0.546 & 0.373 & 0.745 & 0.532 \\
			GAT~\cite{GAT}  & GAT & 0.418 & 0.667 & 0.508 & 0.446 & 0.695 & 0.537 & 0.442 & 0.731 & 0.546 \\
			AliNet~\cite{AliNet}  & GCN+GAT & \underline{0.539} & \underline{0.826} & 0.628 & \underline{0.549} & \underline{0.831} & 0.645 & 0.552 & \underline{0.852} & 0.657 \\
            GCNFlood~\cite{SPA} & GCN & 0.349 & 0.761 & 0.490 & 0.376 & 0.770 & 0.512 & 0.349 & 0.761 & 0.490\\
            DvGNet~\cite{DvGNet} & GAT & 0.534 & \textbf{0.844} & \underline{0.638} & 0.538 & \textbf{0.863} & \underline{0.651} & \underline{0.557} & \textbf{0.881} & \underline{0.668} \\
			\midrule
			decentRL & DAN & \textbf{0.589} & 0.819 & \textbf{0.672}
			& \textbf{0.596} & 0.819 & \textbf{0.678} & \textbf{0.602} & 0.842 & \textbf{0.689} \\
			\bottomrule
	\end{tabular}}
	\label{tab:ea_results}
\end{table*}

\begin{figure*}[t]
	\centering
	\includegraphics[width=.95\textwidth]{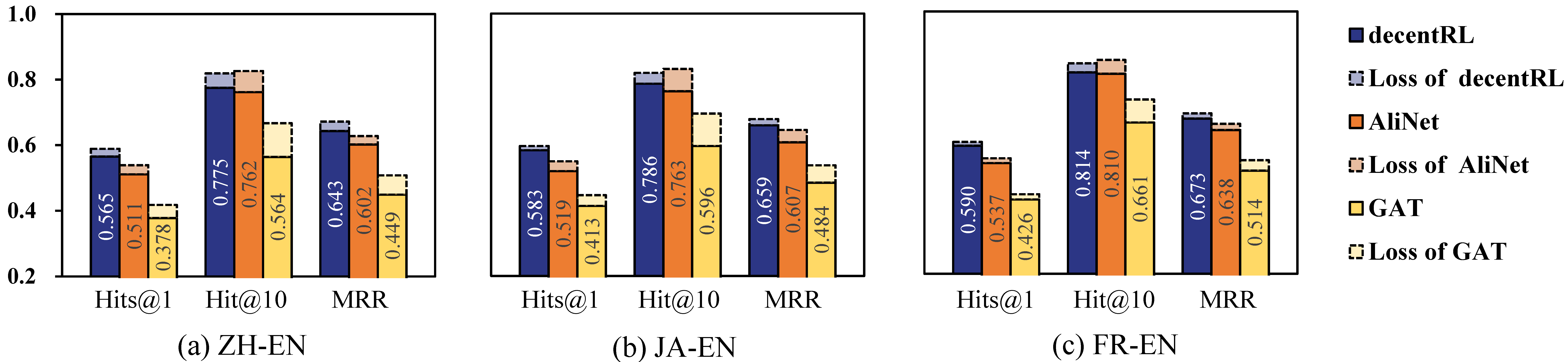}
	\caption{Entity alignment results on open-world DBP15K. Bars with dotted lines denote the performance drop compared with the corresponding results on the conventional datasets.}
	\label{fig:expr_openea}
\end{figure*}

\begin{table}[t]
	\centering
	\caption{Detailed entity alignment results on the open-world ZH-EN dataset.}
	\resizebox{\linewidth}{!}{\scriptsize
		\begin{tabular}{lcccccc}
			\toprule
			\multirow{2}{*}{Method} & \multicolumn{3}{c}{Known Entities} & \multicolumn{3}{c}{Unknown Entities} \\
			\cmidrule(lr){2-4} \cmidrule(lr){5-7}
			& H@1 & H@10 & MRR & H@1 & H@10 & MRR \\ \midrule
			%+LAN transductive 0.096 & 0.362 & 0.182 & 0.059  & 0.289 & 0.133 \\
			AliNet~\cite{AliNet} & \underline{0.535} & \textbf{0.813} & \underline{0.623} & \underline{0.439}  & \underline{0.609} & \underline{0.539} \\
			decentRL & \textbf{0.580} & \underline{0.807} & \textbf{0.660} & \textbf{0.521}  & \textbf{0.679} & \textbf{ 0.592} \\
			\bottomrule
	\end{tabular}}
	\label{tab:detailed_openea}
\end{table}

\section{Experiments}
\label{sec:expr}

We implement the end-to-end framework \emph{decentRL}, and evaluate it on the entity alignment and entity prediction tasks. 
On each task we select the best-performing methods for comparison.
%As the existing methods cannot achieve state-of-the-art results on both tasks, we selected the best-performing methods for comparison on each individual task.

\subsection{Implementation}
\label{app:impl}

\begin{algorithm}[t]
	\caption{Decentralized Knowledge Graph Representation Learning for Entity Alignment}
	\label{alg:decentRL}
	\begin{algorithmic}[1]
		\STATE {\bfseries Input:} The Merged KG $\gG$;
		\STATE Randomly initialize the DAN model $\gM$, entity embeddings $\{\rve|\forall e \in \gE\}$, and relation embeddings $\{\rvr|\forall r \in \gR\}$;
		\REPEAT
		\STATE $\gB \leftarrow \{(e_i, e_j)|(e_i, e_j) \sim \gS\}$ \quad   \textit{// mini-batch training}
		\FOR{$(e_i, e_j) \in \gB$}
		\STATE $\rvg_i, \rve_i \leftarrow \gM(e_i, N_i)$,\quad $\rvg_j, \rve_j \leftarrow \gM(e_j, N_j)$; 
		\STATE Compute the self-distillation loss (Equation~\ref{eq:full_objective});
		\STATE Compute the entity alignment loss (Equation~\ref{eq:align});
		% \tcp{As we only use the copy value of the input entity embedding, we do not update the parameter of the central entity.}
		\STATE Jointly minimize two losses;
		\ENDFOR
		\UNTIL{the performance does not improve.}
	\end{algorithmic}	
	
\end{algorithm}

We present the training procedure of decentRL for entity alignment in Algorithm~\ref{alg:decentRL}. It is worth noting that decentRL does not rely on additional data such as pretrained KG embeddings or word embeddings. The algorithm first randomly initializes the DAN model, entity embeddings, and relation embeddings. The training process follows conventional mini-batch training, akin to existing methods. In addition to the primary entity alignment loss, the algorithm also incorporates a self-distillation loss to guide DAN in generating desired embeddings. It jointly minimizes two losses in each batch until the performance ceases to improve on the validation dataset.

We leverage a contrastive loss \cite{GCN-Align,AliNet} as the primal entity alignment loss to handle aligned entity pairs $\mathcal{S}$, expressed as:
\begin{align}
	\label{eq:align}
	\mathcal{L}_{a} &= \sum_{(i,j) \in \mathcal{S}^{+}}||\mathbf{g}_i - \mathbf{g}_j|| \nonumber\\
	&+ \sum_{(i',j') \in \mathcal{S}^{-}}\alpha\big[\lambda - || \mathbf{g}_{i'} - \mathbf{g}_{j'}||\big]_+,
\end{align}
where $\mathcal{S}^{+}$, $\mathcal{S}^{-}$ represent the positive entity pair set (i.e., the training set) and sampled negative entity pair set, respectively. The term $||\cdot||$ calculates the L2 distance between two entity embeddings. The hyper-parameters $\alpha$ and $\lambda$ are used to regulate the penalty and margin for negative pairs, respectively.

Similarly, for entity prediction, we leverage a decoder to predict missing entities~\cite{CompGCN}. In our experiments, we employ ComplEx~\cite{ComplEx} and DistMult~\cite{DistMult} as the decoders due to their superior performance without compromising efficiency. We initialize the input entity embeddings, relation embeddings, and weight matrices using the Xavier initializer~\cite{initializer}. Detailed parameter settings can be found in Table~\ref{tab:hyperparameter}.

\subsection{Datasets}
\label{sec:expr_dataset}
We consider various datasets for the entity alignment and entity prediction tasks, along with open-world evaluation:

\paragraph{Entity Alignment} In the entity alignment task, we leverage the widely used DBP15K datasets \cite{EVA,MSNEA,MCLEA,JAPE,RSN,BootEA,MuGNN,Meaformer} in our experiment. These datasets encompass three entity alignment settings, each comprising two linked KGs in different languages. For instance, the ZH-EN dataset involves the alignment between Chinese and English entities, leveraging DBpedia as the KG source~\cite{DBpedia}.

\paragraph{Entity Prediction} In the entity prediction task, we use four prominent datasets: (1) FB15K which is a dataset that has been widely used for many years and includes Freebase entities and their relations~\cite{TransE,ConvE,FreeBase,Node+LinkFeat}; (2) WN18 which is another extensively used dataset comprising entities and relations from WordNet~\cite{TransE,ComplEx,RotatE,TuckER,RSN,WordNet,TransR}; 
%It comprises entities and relations from WordNet~\cite{WordNet}. 
(3) FB15K-237 which is a modified version of FB15K with redundant data eliminated~\cite{Node+LinkFeat}; (4) WN18RR which is a modified version of WN18 with redundant data removed as FB15K-237~\cite{ConvE}.

\paragraph{Open-world Datasets} To evaluate the performance of different methods in an open-world scenario, we restructure the entity alignment and entity prediction benchmarks. This involves sampling 20\% of examples from each original testing set to create new entities. Subsequently, we transfer all triplets associated with these new entities to an auxiliary triplet set, exclusively accessible during the testing phase. Consequently, the final testing set encompasses both known entities and unknown entities, thereby augmenting the challenge for KG embedding methods.

Table~\ref{tab:ea_dataset} provides a comparison between the datasets before and after the re-splitting. Although we sample 20\% of the entities in the testing set as new entities, the number of triplets removed from the training set exceed 20\%. As a result, the new datasets present a greater challenge for all KG embedding methods. Notably, in the open-world FB15K-237, the number of training triplets is significantly reduced.

\subsection{Entity Alignment Results}
Table~\ref{tab:ea_results} presents the results of conventional entity alignment. decentRL achieves state-of-the-art performance, surpassing all others in Hits@1 and MRR. AliNet~\cite{AliNet}, a hybrid method combining GCN and GAT, performs better than the methods solely based on GAT or GCN on many metrics. Nonetheless, across most metrics and datasets, decentRL consistently outperforms AliNet, highlighting the robustness of the proposed decentralized attention mechanism.

Although GCN and GAT are generally regarded as inductive models for graph representation learning, our analysis in previous sections suggests their limited applicability on relational KG embedding. In further validation of this, we compare the performance of decentRL with AliNet and GAT on datasets containing new entities. The existing inductive KG embedding methods, such as LAN \cite{LAN}, are unsuitable for adaptation to this task as they are tailored for entity prediction.

Figure~\ref{fig:expr_openea} shows the experimental results. decentRL outperforms both GAT and AliNet across all metrics. While its performance slightly decreases compared to conventional datasets, the other methods experience even greater performance drops in this context. AliNet also outperforms GAT, as it combines GCN and GAT to aggregate different levels of neighbors. The reduced reliance (with GCN) on self-entity embedding contributes to its more resilient performance on datasets with new entities. We also provide more detailed results on ZH-EN in Table~\ref{tab:detailed_openea}, where decentRL surpasses AliNet by a larger margin for the new entities on all metrics.

\begin{table}[t]
	\centering
	\caption{The conventional entity prediction results on FB15K and WN18.}
	
	\resizebox{\linewidth}{!}{
		\begin{tabular}{lccccccc}
			\toprule
			\multirow{2}{*}{Method} & \multirow{2}{*}{\makecell{Backbone\\models}}                                                       & \multicolumn{3}{c}{FB15K} & \multicolumn{3}{c}{WN18} \\
			\cmidrule(lr){3-5} \cmidrule(lr){6-8} & & Hits@1 & Hits@10 &  MRR   & Hits@1 & Hits@10 &  MRR  \\ \midrule
			TransE~\cite{TransE}  & None & 0.297  &  0.749  & 0.463  & 0.113  &  0.943  & 0.495 \\
			RotatE~\cite{RotatE}  & None & \textbf{0.746}  &  0.884  & \underline{0.797}  & 0.944  &  \underline{0.959}  & \underline{0.949} \\
			TuckER~\cite{TuckER}  & None & 0.741  &  \underline{0.892}  & 0.795  & \textbf{0.949}  &  0.958  & \textbf{0.953} \\
			\midrule
			ConvE~\cite{ConvE}  & CNN                                                                         & 0.558  &  0.831  & 0.657  & 0.935  &  0.956  & 0.943  \\
			RSN~\cite{RSN}  & RNN                                                                        & 0.722  &  0.873  &  0.78  & 0.922  &  0.953  & 0.940\\
			\midrule
			R-GCN~\cite{R-GCN} & GCN & 0.601 & 0.842 & 0.696 & 0.697 & \textbf{0.964} & 0.819 \\
			\midrule
			decentRL & DAN &  \underline{0.745} & \textbf{0.901} & \textbf{0.804} & \underline{0.945} & \underline{0.959} & \underline{0.949}\\ \bottomrule
	\end{tabular}}
	\label{tab:ep_results}
\end{table}

\begin{table}[t]
	\centering
	\caption{The conventional entity prediction results on FB15K-237 and WN18RR.}
	
	\resizebox{\linewidth}{!}{
		\begin{tabular}{lccccccc}
			\toprule
			\multirow{2}{*}{Method} & \multirow{2}{*}{\makecell{Backbone\\models}}                                                       & \multicolumn{3}{c}{FB15K-237}  & \multicolumn{3}{c}{WN18RR}   \\
			\cmidrule(lr){3-5} \cmidrule(lr){6-8} & & Hits@1 &    Hits@10     &  MRR  & Hits@1 & Hits@10 &   MRR     \\ \midrule
			TransE~\cite{TransE} & None  &  -  &     0.465      & 0.294 &   -   &  0.501  &  0.226    \\
			RotatE~\cite{RotatE} & None  & 0.241  &     0.533      & 0.338 & \underline{0.428}  &  \textbf{0.571}  &  \underline{0.476}    \\
			TuckER~\cite{TuckER} & None  & \textbf{0.266}  & \textbf{0.544} & \textbf{0.358} & \textbf{0.443}  &  0.526  &  0.470    \\
			\midrule
			ConvE~\cite{ConvE}  & CNN  & 0.237  &     0.501      & 0.325 & 0.400  &  0.520  &  0.430    \\
			RSN~\cite{RSN}  & RNN & 0.202  &     0.453      & 0.280 &  -  & - & - \\
			
			\midrule
			R-GCN~\cite{R-GCN} & GCN & 0.151 & 0.417 & 0.249 & - & - & -  \\
			CompGCN~\cite{CompGCN} & GCN & \underline{0.264}  &     \underline{0.535}      & \underline{0.355} & \textbf{0.443}  &  \underline{0.546}  &  \textbf{0.479}    \\
            NoGE~\cite{NoGE} & GCN & 0.235 & 0.511 & 0.326 & 0.067 & 0.470 & 0.226 \\
            MRGAT~\cite{MRGAT} & GAT & 0.240 & 0.501 & 0.327 & 0.343 & 0.514 & 0.404\\
            CLGAT~\cite{CLGAT} & GAT & 0.255 & 0.530 & 0.348 & 0.409 & 0.535 & 0.450\\
            GATH~\cite{GATH} & GAT & 0.253 & 0.527 & 0.344 & 0.426 & 0.537 & 0.463\\
			\midrule
			decentRL & DAN & 0.261 & \textbf{0.544} & 0.354 & 0.422 & 0.533 & 0.458  \\ \bottomrule
	\end{tabular}}
	\label{tab:ep_results2}
\end{table}

\begin{table}[t]
	\centering
	\caption{Results of employing different decoders on open-world FB15K-237.}
	\resizebox{\linewidth}{!}{\scriptsize
		\begin{tabular}{lcccccc}
			\toprule
			\multirow{2}{*}{Method} & \multicolumn{3}{c}{TransE} & \multicolumn{3}{c}{DistMult} \\
			\cmidrule(lr){2-4} \cmidrule(lr){5-7}
			& H@1 & H@10 & MRR & H@1 & H@10 & MRR \\ \midrule
			+\,LAN~\cite{LAN} & 0.088 & 0.322 & 0.164 & 0.062  & 0.262 & 0.126 \\
			+\,CompGCN~\cite{CompGCN} & \underline{0.144} & \underline{0.336} & \underline{0.211} & \underline{0.160}  & \textbf{0.404} & \underline{0.241} \\
			\midrule
			+\,decentRL & \textbf{0.172} & \textbf{0.383} & \textbf{0.255}
			& \textbf{0.176} & \underline{0.383} & \textbf{0.264} \\
			\bottomrule
	\end{tabular}}
	\label{tab:open_ep_results}
\end{table}

\begin{table}[t]
	\centering
	\caption{Results of employing different decoders on FB15K-237.}
	\resizebox{\linewidth}{!}{
		\begin{tabular}{lcrccrc}
			\toprule
			\multirow{2}{*}{Method} & \multicolumn{3}{c}{TransE} & \multicolumn{3}{c}{DistMult} \\
			\cmidrule(lr){2-4} \cmidrule(lr){5-7}
			& Hits@10 & MR & MRR & Hits@10 & MR & MRR \\ \midrule
			Raw & 0.465 & 357 & 0.294 & 0.419 & 354 & 0.241 \\
			+\,D-GCN~\cite{D-GCN} & 0.469 & 351 & 0.299 & 0.497 & 225 & 0.321 \\
			+\,R-GCN~\cite{R-GCN} & 0.443 & 325 & 0.281 & 0.499 & 230 & 0.324 \\
			+\,W-GCN~\cite{W-GCN} & 0.444 & 1,520 & 0.267 & 0.504 & 229 & 0.324 \\
			+\,CompGCN~\cite{CompGCN} & \underline{0.515} & \underline{233} & \textbf{0.337} & \underline{0.518}  & \underline{200} & \underline{0.338} \\
			\midrule
			+\,decentRL & \textbf{0.521} & \textbf{159} & \underline{0.334}
			& \textbf{0.541} & \textbf{151} & \textbf{0.350} \\
			\bottomrule
	\end{tabular}}
	\label{tab:ep_decoders}
\end{table}

\begin{figure}[!t]
	\includegraphics[width=\linewidth]{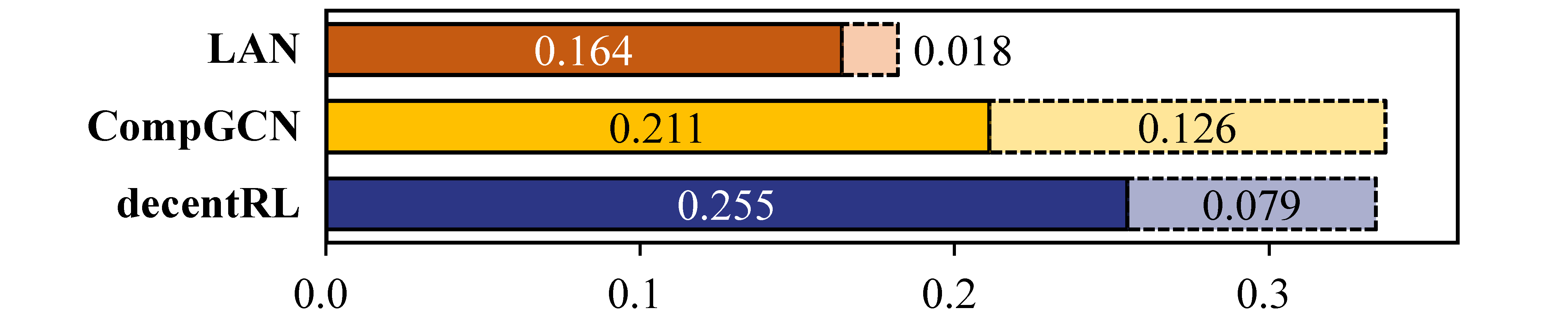}
	\caption{MRR results on open-world FB15K-237, with TransE as the decoder.}
	\label{fig:open_ep}
\end{figure}

\begin{table*}[t]
	\centering
	\caption{Ablation study on the conventional/open-world entity alignment datasets.}
	\resizebox{\textwidth}{!}{
		\begin{tabular}{lccccccccc}
			\toprule
			\multirow{2}{*}{Method} & \multicolumn{3}{c}{{ZH-EN}} & \multicolumn{3}{c}{{JA-EN}} & \multicolumn{3}{c}{{FR-EN}}  \\
			\cmidrule(lr){2-4} \cmidrule(lr){5-7} \cmidrule(lr){8-10}
			& Hits@1 & Hits@10 & MRR & Hits@1 & Hits@10 & MRR & Hits@1 & Hits@10 & MRR \\ \midrule
			decentRL & \textbf{0.589/0.565} & \textbf{0.819/0.775} & \textbf{0.672/0.643}
			& \textbf{0.596/0.583} & \textbf{0.819/0.786} & \textbf{0.678/0.659} & \textbf{0.602/0.590} & \textbf{0.842/0.814} & \textbf{0.689/0.673} \\
			decentRL w/ infoNCE & \underline{0.579}/\underline{0.557} & \underline{0.816}/\textbf{0.775} & \underline{0.665}/\underline{0.637} & \underline{0.591}/\underline{0.574} & \underline{0.816}/\underline{0.785} & \underline{0.673}/\underline{0.652} & \underline{0.593}/\underline{0.583} & 0.834/\underline{0.811} & \underline{0.682}/\underline{0.666} \\
			decentRL w/ L2 & 0.571/0.552 & 0.802/\underline{0.770} & 0.655/0.632 & 0.589/\underline{0.574} & 0.807/0.782 & 0.669/0.650 & 0.591/0.581 & 0.831/0.806 & 0.679/0.664 \\ 
            decentRL w/ self-entity & 0.579/0.551 & 0.812/0.765 & 0.663/0.629 & 0.589/0.573 & 0.812/0.776 & 0.671/0.648 & \underline{0.593}/0.578 & \underline{0.836}/0.806 & 0.681/0.662 \\
			\bottomrule
	\end{tabular}}
	\label{tab:ablation_study}
\end{table*}

\begin{figure*}[t]
	\centering
	\includegraphics[width=\textwidth]{ratio.png}
	% \vspace{-1.2em}
	\caption{The Hits@1, Hits@10, and MRR results on the open-world ZH-EN dataset, in terms of the proportions of unseen entities.}
	\label{fig:ratio}
	%	\vspace{-.5em}
\end{figure*}

\begin{figure}[t]
	\centering
	\includegraphics[width=.9\linewidth]{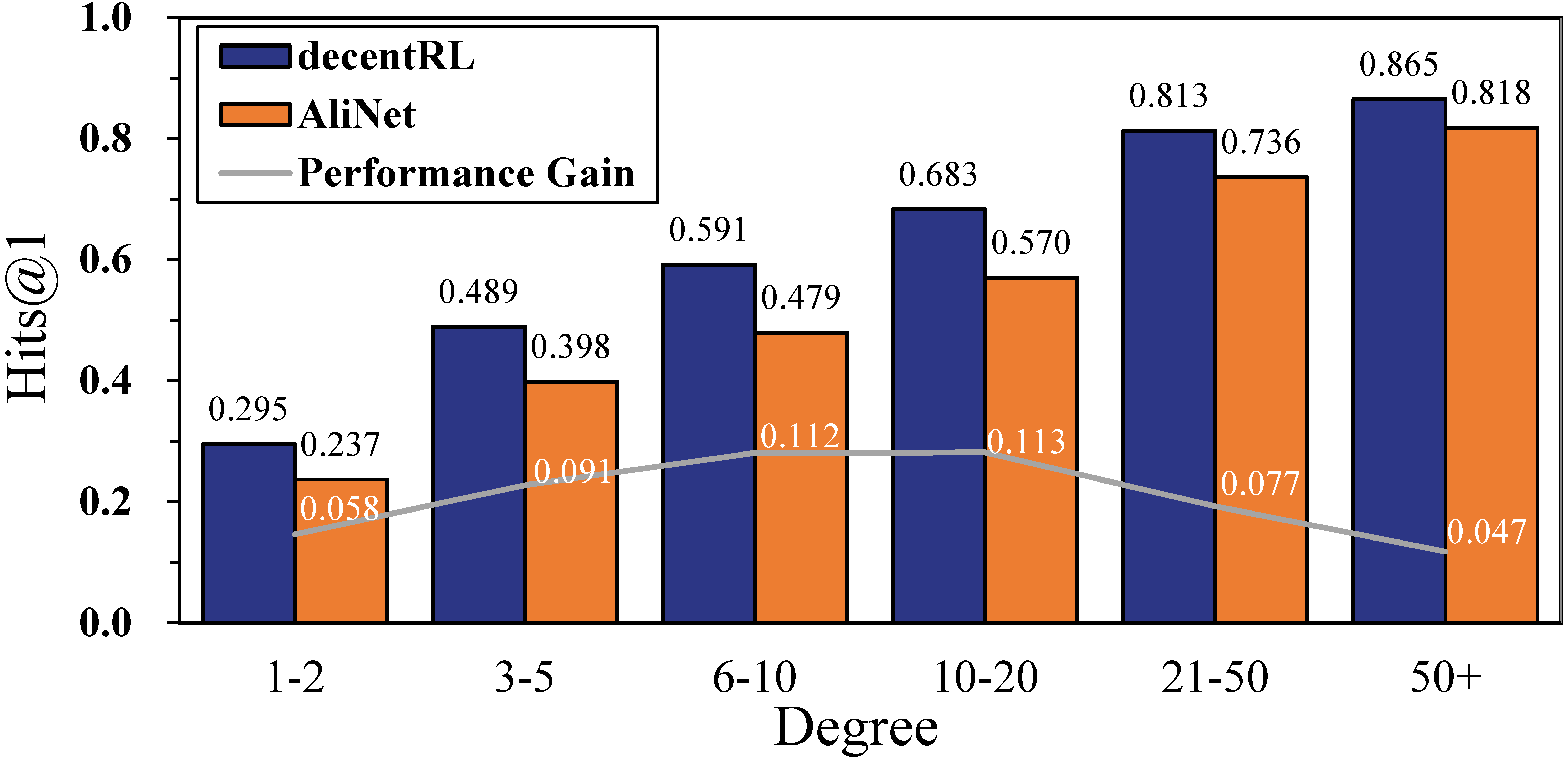}
	\caption{The MRR results in terms of entity degrees, on the ZH-EN dataset.}
	\label{fig:degree_gain}
\end{figure}

\subsection{Entity Prediction Results}

Table~\ref{tab:ep_results} and Table~\ref{tab:ep_results2} present the results for conventional entity prediction. decentRL demonstrates competitive or even superior performance when compared to state-of-the-art methods on the FB15K and WN18 benchmarks, showcasing its efficacy in entity prediction. While on the FB15K-237 and WN18RR datasets, the performance of decentRL is slightly below the best-performing methods, it does achieve the best Hits@10 on FB15K-237. It is worth noting that FB15K-237 and WN18RR pose greater challenges for methods not tailored to this task, such as RSN~\cite{RSN} and decentRL.

We further evaluate decentRL alongside the open-world KG embedding method LAN~\cite{LAN} and the best-performing GNN-based method CompGCN~\cite{CompGCN} on FB15K-237 with new entities. As shown in Figure~\ref{fig:open_ep}, decentRL significantly outperforms CompGCN in this setting. LAN, specifically designed for new entities, experiences minimal performance loss, but faces challenges with known entities. Its MRR under the conventional setting (0.182) is markedly lower than that of CompGCN (0.355) and decentRL (0.354), even falling below the performance of the classical KG embedding method TransE (0.294).

In Table~\ref{tab:open_ep_results}, we present more detailed entity prediction results on open-world FB15K-237, considering the influence of different decoders. Our observations indicate that decentRL consistently outperforms the other methods across most metrics when using TransE and DistMult as decoders. Furthermore, we provide results on the conventional FB15K-237 in Table~\ref{tab:ep_decoders}, affirming similar conclusions.

\section{Further Analysis}
In this section, we conduct experiments to analyze the proposed method in more detail.

\subsection{Ablation Study}

In order to evaluate the efficacy of each module, we implement several alternative methods: decentRL w/ infoNCE, decentRL w/ L2 denote the versions where we replace self-distillation with InfoNCE~\cite{SSL} and L2, respectively. decentRL w/ self-entity denotes the version involving self-entity. 

The results in Table \ref{tab:ablation_study} demonstrate that all variants of decentRL achieves state-of-the-art performance on Hits@1, empirically proving the superiority of using neighbor context as the query vector for aggregating neighbor embeddings. The proposed decentRL outperforms both decentRL w/ infoNCE and decentRL w/ L2, providing empirical evidence for the effectiveness of self-distillation. It is also worth noting that, decentRL w/ self-entity exhibits a significant performance drop on datasets containing new entities. In this version, the input entities themselves participate in their own aggregations, potentially disrupting the roles of the input embeddings. The lower performance on the open-world datasets, compared to other variants, aligns with the assumptions discussed in previous sections.

\begin{table*}[t]
	\centering
	\caption{Performance of decentRL with different embedding-sizes.}
	
	\resizebox{.7\textwidth}{!}{
		\begin{tabular}{ccclcclccl}
			\toprule
			\multirow{2}{*}{Embedding-size} & \multicolumn{3}{c}{{ZH-EN}} & \multicolumn{3}{c}{{JA-EN}} & \multicolumn{3}{c}{{FR-EN}}  \\
			\cmidrule(lr){2-4} \cmidrule(lr){5-7} \cmidrule(lr){8-10}
			& Hits@1 & Hits@10 & MRR & Hits@1 & Hits@10 & MRR & Hits@1 & Hits@10 & MRR \\ \midrule
			64 & 0.429 & 0.644 & 0.508 & 0.474 & 0.673 & 0.547 & 0.468 & 0.704 & 0.554 \\
			128 & 0.511 & 0.726 & 0.590 & 0.541 & 0.745 & 0.617 & 0.535 & 0.773 & 0.623 \\ 
			256 & \underline{0.560} & \underline{0.785} & \underline{0.643} & \underline{0.578} & \underline{0.791} & \underline{0.657} & \underline{0.578} & \underline{0.817} & \underline{0.665} \\
			512 & \textbf{0.589} & \textbf{0.819} & \textbf{0.672} & \textbf{0.596} & \textbf{0.819} & \textbf{0.678} & \textbf{0.602} & \textbf{0.842} & \textbf{0.689} \\
			\bottomrule
	\end{tabular}}
	\label{tab:para_efficiency}
\end{table*}

\begin{figure*}[t]
	\centering
	\includegraphics[width=.95\textwidth]{expr_eachlayer.png}
	% \vspace{-1.2em}
	\caption{The Hits@1 results of each layer and the concatenation. The results of AliNet are from \cite{AliNet}. It has no Layer 3 and Layer 4 scores as the best performance is achieved by a two-layer model.}
	\label{fig:expr_eachlayer}
	%	\vspace{-.5em}
\end{figure*}

\begin{table}[t]
\vspace{-1em}
	\centering
	\caption{Training time (seconds) on ZH-EN.}
	\resizebox{.9\linewidth}{!}{\scriptsize
		\begin{tabular}{crrrr}
			\toprule
			\multirow{2}{*}{Hidden-size} & \multicolumn{2}{c}{decentRL} & \multicolumn{2}{c}{AliNet}  \\
			\cmidrule(lr){2-3} \cmidrule(lr){4-5}  
			& per epoch & total & per epoch & total \\ \midrule
			64  & 1.6 & \textbf{137.7} & \textbf{1.2} & 1,317.6   \\
			128 & 2.0 & \textbf{138.2} & \textbf{1.4} & 1,301.2   \\
			256 & 2.2 & \textbf{161.4} & \textbf{1.8} & 1,297.8   \\
			512 & \textbf{2.4} & \textbf{236.3} & \textbf{2.4}
			& 1,346.5  \\
			\bottomrule
	\end{tabular}}
	% \vspace{-1em}
	\label{tab:running_time}
\end{table}

\subsection{An Analysis of Proportions of Unseen Entities}
We conduct experiments to explore the impact of the numbers of unseen entities on the performance in open-world entity alignment. We present the results on the ZH-EN datasets in Figure~\ref{fig:ratio}. Clearly, the performance gain achieved by leveraging our method significantly increases when there are more unseen entities. For example, when only 20\% of entities are unseen, decentRL outperforms AliNet on Hits@1 by 9.2\%, while this margin extends to 35.9\% when 80\% of entities are unseen. Overall, decentRL demonstrates significant advantages as new entities are added to KGs.

\subsection{An Analysis of Performance Gain and Entity Degrees}
We conduct experiments to investigate the performance gain concerning entity degrees. Typically, an entity with a higher degree indicates that it has more neighboring entities. Consequently, the computation of attention scores to aggregate these neighbors becomes crucial.

The results on the ZH-EN dataset are depicted in Figure~\ref{fig:degree_gain}. For entities with only a few neighbors, the advantage of leveraging DAN is not significant. However, as the degree increases, incorporating DAN yields more performance gain. This upward trend halts until the degree exceeds 20. Overall, DAN exhibits significantly better performance than GCN, GAT, or their combination. The decentralized attention, which considers neighbors as queries, consistently outperforms the centralized GAT across varying entity degrees.

\subsection{An Analysis of Parameter Efficiency}

In Table~\ref{tab:para_efficiency}, we present the entity alignment results concerning various embedding-sizes. Our findings indicate that decentRL with an embedding size of 256 is adequate to outperform AliNet with an embedding size of 512. Furthermore, even a modestly sized decentRL (with an embedding size of 64) can surpass vanilla GAT with an embedding size of 512.

\subsection{An Analysis of Computational Cost}

We conduct an analysis of the training time for decentRL and AliNet with varying hidden-sizes on a V100 GPU, as detailed in Table~\ref{tab:running_time}. We employ a two-layer AliNet (each layer comprising one GCN and one GAT) and a four-layer decentRL. The two methods exhibit comparable running times per epoch. AliNet runs marginally faster than decentRL with smaller hidden sizes, but the total training time of decentRL is notably less than that of AliNet. This difference can be attributed to the faster convergence of decentRL (averaging 55 epochs) compared to AliNet (averaging 280 epochs), potentially benefiting from self-distillation.

\subsection{An Analysis of Multi-layer Models}
We conduct experiments to validate our assumption regarding the role of input embeddings and their correlation with output embeddings. The Hits@1 results of decentRL and AliNet at each layer are depicted in Figure~\ref{fig:expr_eachlayer}. 

The performance of decentRL at the input layer notably lags behind that of other layers and AliNet. As discussed in previous sections, decentRL does not use the embedding of the central entity as input when generating its output embedding. However, this input embedding can still accumulate knowledge by participating in the aggregations of its neighbors. The acquired information may not necessarily reside in the same dimension for a pair of aligned entities at this layer, which accounts for the comparatively lower performance of this layer. Nevertheless, it still contains useful information for entity alignment. Additionally, decentRL benefits from concatenating the embeddings from multiple layers. The optimal performance is achieved by a four-layer decentRL.

\section{Conclusion}
In this work, we propose Decentralized Attention Network for knowledge graph embedding and introduce self-distillation to enhance its ability to generate desired embeddings for both known and unknown entities. We provide theoretical justification for the effectiveness of our proposed learning paradigm and conduct comprehensive experiments to evaluate its performance on entity alignment and entity prediction, considering scenarios with and without new entities. Our experimental results demonstrate state-of-the-art performance of the proposed method on conventional and open-world benchmarks for both entity alignment and entity prediction tasks. Our method not only provides a solution for knowledge graph representation learning but also offers valuable insights into the potential of decentralized attention mechanisms for other graph-based applications.

\section*{Acknowledgment}
This work is funded by National Natural Science Foundation of China (NSFCU23B2055/NSFCU19B2027/NSFC91846204), Zhejiang Provincial Natural Science Foundation of China (No.LGG22F030011), and Fundamental Research Funds for the Central Universities (226-2023-00138).

\appendix
\section{Proofs of Things}
\subsection{Proof of Proposition~\ref{prop:lower-bound}}
\label{proof:low-bound}
\begin{proof}
	Recall the second term in Equation~(\ref{eq:full_objective}):
	\begin{align}
		\label{eq:former_objective}
		\mathop{\text{argmax}}_{\mathbf{g}_i, f} \mathop{\mathbb{E}}_{X_i} \log\Big( \frac{f(\mathbf{g}_i, \hat{\mathbf{e}}_i)}{ \sum_{e_j \in X_i} f(\mathbf{g}_i, \hat{\mathbf{e}}_j)} \Big).
	\end{align}
	Here, $\mathbf{e}_i$ serves as a ``pre-trained'' teacher model to instruct the student $\rvg_i$. The learnable parameters are $\mathbf{g}_i$ and $f$. To elaborate, we can expand this term as follows:
	\begin{align}
		\label{eq:expand_objective}
		(\mathbf{N}_i ,\varTheta, f)= \mathop{\text{argmax}}_{\mathbf{N}_i ,\varTheta, f} \mathop{\mathbb{E}}_{X_i} \log\Big( \frac{f(G(\mathbf{N}_i), \hat{\mathbf{e}}_i)}{ \sum_{e_j \in X_i} f(G(\mathbf{N}_i), \hat{\mathbf{e}}_j)} \Big),
	\end{align}
	where $\mathbf{N}_i = \{\mathbf{e}_j| e_j \in N_i \}$ represents the input neighbor embedding set for $\mathbf{e}_i$ and $\varTheta$ denotes the parameters of DAN. Since the optimal $\varTheta$ depends on the neighbor embedding set $\mathbf{N}_i$, and the optimal density function $f$ also relies on the output of DAN, exhaustive search across all spaces to find the best parameters is infeasible. Therefore, we opt to optimize a weaker lower-bound on the mutual information $I(\mathbf{g}_i, \hat{\mathbf{e}}_i)$~\cite{CRD}. In this case, a relatively optimal neighbor embedding $\mathbf{e}_x^*$ of $e_i$ in Equation~(\ref{eq:expand_objective}) is:
	\begin{align}
		\label{eq:optimal_neighbor}
		\mathbf{e}_x^* = \mathop{\text{argmax}}_{\mathbf{e}_x} \mathop{\mathbb{E}}_{X_i} \log\Big( \frac{f(G(\mathbf{N}_i), \hat{\mathbf{e}}_i)}{ \sum_{e_j \in X_i} f(G(\mathbf{N}_i), \hat{\mathbf{e}}_j)} \Big),
	\end{align}
	and we then get:
	\begin{align}
		\label{eq:lower-bound}
		\widehat{I}(\mathbf{g}_i, \hat{\mathbf{e}}_i | \mathbf{e}_x^*) &= \mathop{\mathbb{E}}_{X_i} \log\Big( \frac{f(G(\{\ldots,\mathbf{e}_x^*,\ldots\}, \hat{\mathbf{e}}_i)}{ \sum_{e_j \in X_i} f(G(\{\ldots,\mathbf{e}_x^*,\ldots\}, \hat{\mathbf{e}}_j)} \Big)\\
		&\leq \mathop{\mathbb{E}}_{X_i} \log\Big( \frac{f^*(G^*(\mathbf{N}_i^*), \hat{\mathbf{e}}_i)}{ \sum_{e_j \in X_i} f^*(G^*(\mathbf{N}_i^*), \hat{\mathbf{e}}_j)} \Big) \\
		&= \widehat{I}(\mathbf{g}_i^*, \hat{\mathbf{e}}_i) \label{eq:infoNce1}\\
		&\leq \widehat{I}(\mathbf{g}_i^*, \hat{\mathbf{e}}_i) + \log(|X_i|)\label{eq:infoNce2}\\
		&\leq I(\mathbf{g}_i, \hat{\mathbf{e}}_i) = I(\mathbf{g}_i, \mathbf{e}_i),
	\end{align}
	where the notation $^*$ denotes the optimal setting for the corresponding parameters. Equations~(\ref{eq:infoNce1}) and (\ref{eq:infoNce2}) are the conclusion of InfoNCE, assuming that $|X_i|$ is sufficiently large. The aforementioned equations imply that optimizing a single neighbor embedding $\mathbf{e}_x$ also provides a lower bound on the mutual information without the necessity of perfectly assigning other parameters.
\end{proof}

\subsection{Proof of Proposition~\ref{prop:auto-distillation}}
\label{proof:auto-distillation}
\begin{proof}
	Suppose that the neighbor entity $e_x$ of $e_i$ has more than one neighbor, we can collectively optimize those cases to obtain $\rve_x^*$:
	\begin{align}
		\label{eq:auto-distiller}
		\mathbf{e}_x^* &= \mathop{\text{argmax}}_{\mathbf{e}_x} \sum_{e_j\in N_x} \mathop{\mathbb{E}}_{X_j} \log\Big( \frac{f(G(\mathbf{N}_j), \hat{\mathbf{e}}_j)}{ \sum_{e_k \in X_j} f(G(\mathbf{N}_j), \hat{\mathbf{e}}_k)} \Big)
	\end{align}
	Note that the above equation is identical to the first term of Equation~(\ref{eq:full_objective}), i.e.,
	\begin{align}
		\label{eq:latter_objective}
		\mathop{\text{argmax}}_{\mathbf{e}_i} \sum_{e_j\in N_i} \mathop{\mathbb{E}}_{X_j} \log\Big( \frac{f(\mathbf{g}_j, \hat{\mathbf{e}}_j)}{ \sum_{e_k \in X_j} f(\mathbf{g}_j, \hat{\mathbf{e}}_k)} \Big).
	\end{align} 
	This implies optimizing $\rvg_i$ in Equation~(\ref{eq:former_objective}) consequently contributes to optimizing the input embedding of $e_i$'s neighbors (e.g., $\rve_x$), and verse visa. In other words, there is no explicit need to optimize $\rve_i$ separately because it can be optimized during the learning $\rvg_j$, where $e_j$ is one of the neighbors of $e_i$.
\end{proof}

\bibliography{references_with_page_number}
\bibliographystyle{elsarticle-num}

\end{document}